%% file: MNSBMarxiv.tex
\documentclass[reprint,amsmath,amssymb,mathtools,aps,pre]{revtex4-1}
\pdfoutput=1
\usepackage{amsfonts,bbm}
\usepackage{times}
\usepackage{graphicx}
\usepackage{color}
\usepackage{psfrag}
\usepackage{bbm}
\usepackage{latexsym}
\usepackage{url}
\usepackage{bm}
\usepackage[caption=false]{subfig}
\usepackage{enumitem}

\usepackage{xspace}
\newcommand{\nnspace}{\!\!\!}

\newcommand{\CRP}{\mathcal{C}}

\newcommand{\Poisson}{\mathcal{P}}
\newcommand{\Gam}{\mathcal{G}}
\newcommand{\m}[1]{\bm{#1}}

\newcommand{\Multinomial}{\mathcal{M}}

\newcommand{\Aobs}{A^*}

\newcommand{\est}[1]{\ensuremath{\hat{#1}}\xspace}

\newcommand{\figref}[1]{FIG.~\ref{#1}}

\newcommand{\bigO}[1]{\ensuremath{\mathcal{O}(#1)}\xspace}

\begin{document}

\title{Efficient inference of overlapping communities in complex networks}

\author{Bjarne {\O}rum Fruergaard}
\email{bowa@dtu.dk}
\altaffiliation[Also at ]{Adform ApS, DK-1103 K{\o}benhavn K, Denmark}

\author{Tue Herlau}
\email{tuhe@dtu.dk}
\affiliation{%
	Section for Cognitive Systems, DTU Compute, Technical University of Denmark, DK-2800 Kongens Lyngby, Denmark
}%

\date{\today}

\begin{abstract}
\input{abstract}

\end{abstract}


\maketitle


\section{\label{section:intro}Introduction}
\input{introduction}

\section{\label{section:model}Methods}
\input{methods}

\section{\label{section:results}Results and discussion}
\input{results}

\section{\label{section:conclusion}Conclusion}
\input{conclusion}

\bibliography{Library}

\end{document}

%% file: abstract.tex
We discuss two views on extending existing methods for complex network modeling which we dub the \textit{communities first} and the \textit{networks first} view, respectively. Inspired by the networks first view that we attribute to \citeauthor{white1976social}\cite{white1976social}, we formulate the multiple-networks stochastic blockmodel (MNSBM), which seeks to \textit{separate} the observed network into subnetworks of different types and where the problem of inferring structure in each subnetwork becomes easier. We show how this model is specified in a generative Bayesian framework where parameters can be inferred efficiently using Gibbs sampling. The result is an effective multiple-membership model without the drawbacks of introducing complex definitions of "groups" and how they interact. We demonstrate results on the recovery of planted structure in synthetic networks and show very encouraging results on link prediction performances using multiple-networks models on a number of real-world network data sets.

%% file: introduction.tex
An important theme in modern network science is the inference of structure in complex networks. The simplest and most well studied type of structure is based on partitions of vertices commonly denoted as \emph{blockmodels}. The basic assumption in block modeling is that the vertices are partitioned into non-overlapping sets, the \emph{blocks}, and the probability of observing an edge between two vertices depends only on the block each vertex belong to. This implies vertices in the same block are structurally equivalent~\cite{white1976social}.
 If the partition of vertices into blocks and the other parameters in the model are all considered as random variables in a Bayesian framework the resulting method is commonly known as the \emph{stochastic blockmodel} (SBM)~\cite{holland1983stochastic}.

The SBM has two desirable properties. Firstly, it is sufficiently flexible to capture many different patterns of interaction. Secondly, the model is easy to implement and allows inference of structure in larger networks.
When considering extensions of the SBM the following line of reasoning is often followed: The SBM makes use of a latent structure where the vertices are divided into groups or communities. The assumption that everything belongs to exactly one group (friends, family, coworkers, etc.) is too simplistic since these communities often overlap in reality, hence the assumption each vertex belong to one group should be relaxed. This line of thinking lead to two classes of models depending on how the partition-assumption is relaxed. The first type replaces the partition structure of the vertices with a multi-set, that is, a collection of non-empty but potentially overlapping subsets of the vertices~\cite{miller2009nonparametric,palla2012infinite,morup2011infinite}. The second is a continuous relaxation where for each vertex and each "community" there is a continuous parameter specifying the degree of which the vertex is associated with the community~\cite{airoldi2008mixed,PhysRevE.84.036103,menon2011link}. A difficulty with both approaches is that when the assumption of each vertex belonging to a single block is relaxed, the probability that two vertices link to each other must be specified as a function of all the blocks the two vertices are associated with or belong to. The multitude of ingenious ways this problem is solved attests to this being a difficult problem and many of these methods are difficult to implement efficiently and are therefore only applied to small networks.

We wish to emphasize that the above mentioned approaches to extend the basic SBM to models with overlapping blocks, both derive from a more basic assumption; namely that the main goal of block modeling is to model groups or communities of vertices. We dub this view the \emph{communities first} view to emphasize the focus on detecting latent group structure in the network. Comparing to the original work on block modeling by \citeauthor{white1976social}\citep{white1976social}, we argue there are two subtle but important distinctions from this more modern interpretation of block modeling: Firstly, that the block only exists as a postulate of structural equivalence between vertices and are specifically not thought to have an interpretation as communities. Secondly, this partitioning of vertices into blocks is only admissible by carefully keeping edges of different \emph{types} as distinct networks. That is to say, that by representing edges of different types as distinct networks, the \emph{simplifying} assumption of stochastic equivalence across blocks becomes permissible. To emphasize this distinction we call this view the \emph{networks first} view.

In this work we propose a method which focuses on the networks first view. We consider a single network as being composed of multiple networks and the principled goal of network modeling is to de-mix this structure into separate networks and model each of the networks with a simpler model. In our work we consider this simplified model to be a stochastic blockmodel, however we emphasize the idea naturally extents to many other types of latent structure including models of overlapping structure. The resulting model, which we name the \emph{multi-network stochastic blockmodel} (MNSBM), has several benefits
\begin{enumerate}[nosep, label={\roman*}]
	\item Our sampler for Bayesian inference is easy to specify, the hardest part boiling down to a discrete sampling problem, which can be efficiently parallelized.
	\item The inference is nonparametric in that it infers the model order automatically for each subnetwork.
	\item The method is easily extended to include hybrid models such as models of overlapping hierarchies and blockmodels.
\end{enumerate}
The remainder of the paper is organized as follows: In section~\ref{section:white} we will argue more carefully that the networks first view is found in the original work by \citeauthor{white1976social}, in section~\ref{section:model} we will introduce the MNSBM and in section~\ref{section:results} demonstrate our model is able to de-mix planted structure in synthetic data as well as successfully increase link prediction performance on real-world data sets by modeling networks as multiple, overlapping SBMs. 



\subsection{Assumptions of block modeling}\label{section:white}

\citeauthor{white1976social} considers models for multiple networks defined on the same set of vertices. An example is the Samson monastery networks dataset~\cite{sampson1969crisis} in which there are 8 networks where each network is comprised by the answer of the monks to specific question such as the degree to which they like, praise, esteem, etc., other monks. In the terminology of the article each of these networks represents a \emph{type} and an edge in a particular network is an edge of that type. For instance an edge between two monks can be of the type "like" or "praise" etc. The distinction into multiple networks is taken as fundamental:
\begin{quote}
We take as given the incidence of each of several distinct types of tie (...). Each is a separate network to be contrasted with other such networks, rather than merged with them to form a complex bond between each pair of actors. This analytic segregation of network types is basic to our framework~\citep[p731]{white1976social}
\end{quote}
It is worth emphasizing why according to \citeauthor{white1976social} the segregation into different networks is considered basic. The blockmodel hypothesis is given as five points, and we pay special attention to the following two: 
\begin{quote}
First, structural equivalence requires that members of the population be partitioned into distinct sets, each treated homogeneously not only in its internal relations but also in its relations to each other such set. (...) Third, many different types of tie [edges] are needed to portray the social structure [i.e.~communities] of a population.\citep[p739]{white1976social}
\end{quote}
However if a block is simply defined as structurally equivalent vertices (as opposed to a community of vertices), it is on this definition no longer obvious what it means for two blocks to overlap since the overlap would break structural equivalence. The de-emphasis on blocks as capturing explicit group structure and emphasis on the need for types of ties lead us to dub this the \textit{networks first view}. 

It is worth contrasting this with a modern view on blockmodels where blocks are taken to signify structure. For instance \citeauthor{latouche2011}~\cite{latouche2011} first discuss the blockmodel as introduced by \citeauthor{white1976social} and then discuss the point of contention:
\begin{quote}
A drawback of existing graph clustering techniques is that they all partition
the vertices into disjoint clusters, while lots of objects in real world applications
typically belong to multiple groups or communities.~\citep[p310]{latouche2011}
\end{quote}
thus a block or group in this view clearly reflects a real entity thought to exist in the graph, which vertices may or may not belong to, and not simply structural equivalence; hence the term \textit{communities first view}. We wish to emphasize that we do not consider the communities first view on network modeling as being wrong or mistaken, for instance \citeauthor{latouche2011} (and references therein) consider many concrete instances where it is thought to hold. However, we consider it a particular hypothesis on the structure of the network composed of the following two assumptions; (i) the network is composed of links of homogeneous type, and (ii) the network should be thought of as containing groups the vertices may be members to and these groups explain the links. On the contrary, the networks first view considers the complexity of the network as primarily being derived from it containing a mixture of networks of many types. For instance when we collect a network of friendships on an online social network, we may suppose the friendships fall into several categories such as ``friendship''-ties, ``family''-ties, ``work colleague''-ties, and so on. It is the overlap of these distinct types of ties that induces much of the difficulty in modeling complex networks, and when these networks are kept separate, the problem of inferring structure simplifies potentially to the point where naive assumptions such as structural equivalence (as in the SBM) may suffice.

\begin{figure*}[t]
	
	\subfloat[][Observed network] { %
		\scalebox{0.8}{ %
			\includegraphics{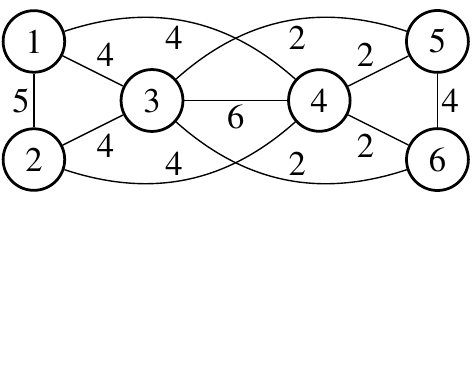}
			\label{fig:ex-obs-network}
		}
	}
	\hfill
	\subfloat[][MNSBM, $S=1$]{ %
		\scalebox{0.8}{
			\includegraphics{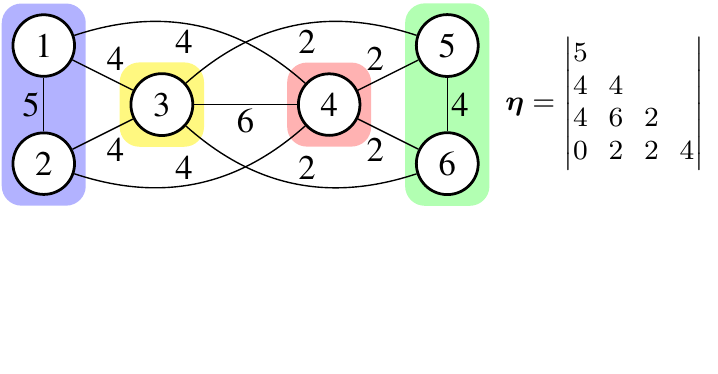}
			\label{fig:ex-disjoint-sbm}
		}
	}
	\hfill
	\subfloat[][MNSBM, $S=2$]{ %
		\scalebox{0.8}{
			\includegraphics{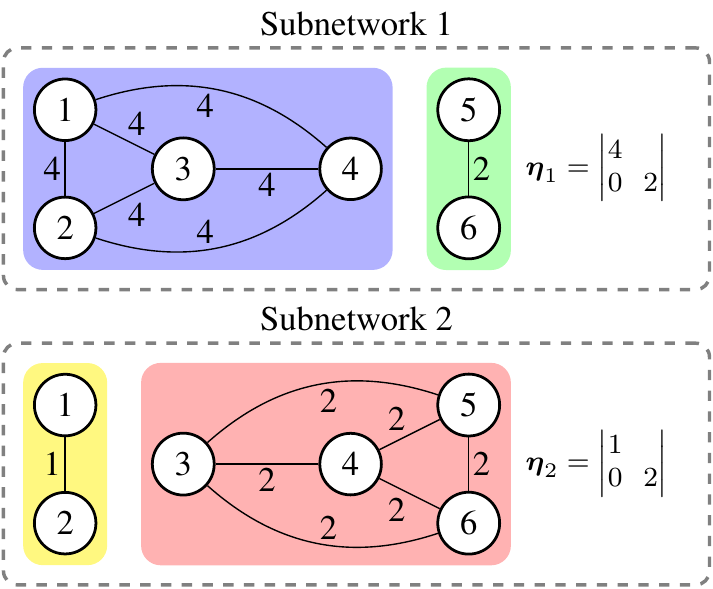}
			\label{fig:ex-mnsbm}
		}
	}
	\caption{(Color online) Examples of networks illustrating the generative models of MNSBM. (a) An example of an undirected multigraph where edge weights represent the number of edges. (b) The same network with a clustering into 4 vertex communities according to MNSBM with a single network, which on average generates the observed network. (c) The same network again, but this time divided in two as per an MNSBM of two networks. The summation of the networks in (c) will also on average generate the observed edges.}
	\label{fig:example-networks}
\end{figure*}

%% file: methods.tex
Consider a network comprised of $n$ vertices, $1,2,\dots,n$, and let $\Aobs_{ij}$ denote the observed number of edges between vertices $i$ and $j$. The reason we allow multiple edges between vertices will be apparent later, however for simplicity (but without loss of generality) we will otherwise consider $\m \Aobs$ as being symmetric and without self-loops, i.e. $\Aobs_{ij} = \Aobs_{ji}$ and $\Aobs_{ii} = 0$. In line with the networks first view we consider $\m \Aobs$ as arising from multiple networks, $\m A^1, \dots, \m A^S$, which have been aggregated due to an unknown data registration process. In the terminology above each network is comprised by a particular type of edge corresponding to for instance different social relations. We will denote the process of aggregation by a function $h$. In principle this could be a function of $S$ networks, however we will make the assumption
\begin{align}
\Aobs_{ij} = h(A_{ij}^1 + A_{ij}^2 + \cdots + A^S_{ij})
\end{align}
and in particular be interested in the case where $h$ is the heavy-side step function $H$, defined as 1 when the input is positive and otherwise zero. This corresponds to the natural assumption; we discover an edge between two vertices if there is an edge in any of the networks of different edge types. Next we assume each network $\m A^s$ arise from a model $\mathcal{M}_s$ with latent parameters $\Phi_s$. In this case
\begin{align}
\Phi^1,\dots,\Phi^S  & \sim P(\cdot) & & \nnspace \textit{not necessarily independently} \\
A^s_{ij} & \sim P(\cdot | \Phi^s) & & \nnspace\textit{independently} 
\end{align}
\begin{align}
& p(\m \Aobs, \m A^1,\dots,\m A^S, \Phi^1,\dots,\Phi^S ) \\ 
& = p(\Phi^1,\dots,\Phi^S)\prod_{i<j} \delta_{\Aobs_{ij} - h(\sum_s A_{ij}^s)}  \prod_s p(A^s_{ij} | \Phi^s) 
\end{align}
From this we can easily extract the marginal probabilities:
\begin{align}
p(A_{ij}^s | \cdots) & \propto p(A^s_{ij} | \Phi^s)  \delta_{\Aobs_{ij} - h(\sum_s A_{ij}^s)}  \\
p(\Phi^s | \cdots) & \propto p(A^s | \Phi^s) p(\Phi^1,\dots,\Phi^S)
\end{align}
An important special case is if the marginal probabilities $p(A_{ij}^s | \Phi^s)$ are Poisson distributed and the parameters $\Phi^1, \dots, \Phi^S$ are independent. We may write this as $p(A_{ij}^s | \Phi^s) = \Poisson(A_{ij}^s | \eta^s_{ij})$ where $\mathcal P$ denotes the Poisson distribution and the rates $\eta_{ij}^s$ are considered part of the parameter vector $\Phi^s$. Recall the following basic properties of Poisson random variables: If $X_1,\dots, X_k$ is a set of $k$ independent Poisson random variables then their sum $X = \sum_i X_i$ is Poisson distributed with rate $\eta = \sum \eta_i$ and the conditional distribution of $(X_1,\dots,X_k)$ on $X = k$ is distributed as a multinomial distribution ($\Multinomial$):
\begin{align}
X_1,\dots,X_k | X = n & \sim \mathcal{M}\left(\  \m \cdot \ | \frac{\eta_1}{\eta_0}, \cdots, \frac{\eta_k}{\eta_0}, n\right)
\end{align}
Introducing $\eta_{ij} = \sum_{s=1}^S \eta_{ij}^s$ then with these assumptions it follows
\begin{align}
p(A_{ij} | \cdots ) & \propto \mathcal{P}(A_{ij} | \eta_{ij})\delta_{ (\Aobs_{ij} - h( A_{ij})) } \label{eq:9}\\
p(A_{ij}^1, \dots, A_{ij}^S | \cdots ) & \propto \mathcal{M}\left(A_{ij}^1,\dots,A_{ij}^S | \frac{\eta_{ij}^1}{\eta_{ij}}, \cdots, \frac{\eta_{ij}^S}{\eta_{ij}}, A_{ij}\right) \label{eq:10} \\
p(\Phi^s | \cdots) & \propto p(A^s | \Phi^s) p(\Phi^s) \label{eq:11}
\end{align}
Since the restriction in eq.~\eqref{eq:9} is on a univariate density it will for all reasonable choices of $h$ be easy to handle analytically. In the particular case of the Heaviside function we have 
\begin{align}
p(A_{ij} | \cdots) & = \left\{ 
\begin{array}{cc}
  \delta_{A_{ij}} & \mbox{if $\Aobs_{ij} = 0$,} \\
  \frac{\mathcal{P}(A_{ij}|\eta_{ij})}{1-\mathcal{P}(0|\eta_{ij})} & \mbox{otherwise.} 
\end{array}\right. 
\end{align}
Accordingly, eqs.~\eqref{eq:9} and \eqref{eq:10} may be sampled very quickly independently for each edge in the observed network $\m A^*$ and sampling eq.~\eqref{eq:11} is only as complex as sampling a single network model and, when considering $S$ networks, these parameters may be sampled independently of each other. 

\begin{figure*}[t]
	\centering
	\begin{tabular*}{\textwidth}{@{\extracolsep{\fill} } ccc}
		$\lambda=0.0$ & $\lambda=0.5$ & $\lambda=1.0$
		\\
		\includegraphics[width=0.3\textwidth]{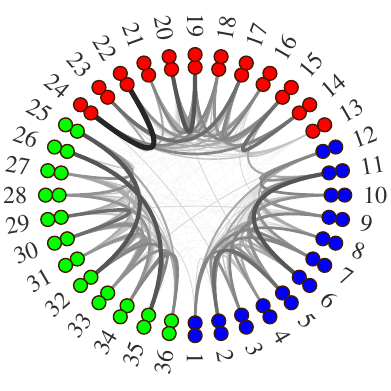}
		&
		\includegraphics[width=0.3\textwidth]{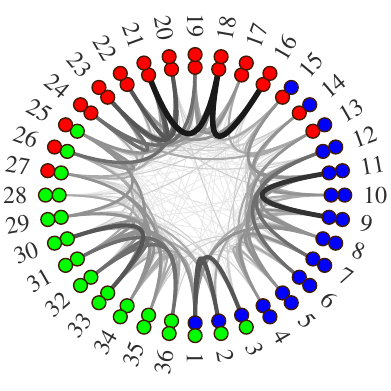}
		&
		\includegraphics[width=0.3\textwidth]{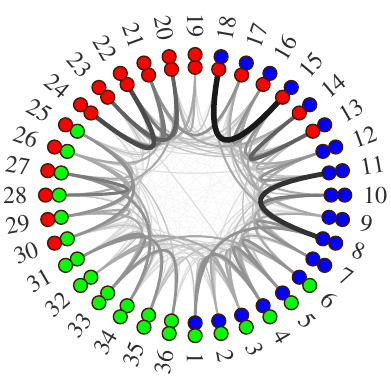}
	\end{tabular*}
	\caption{(Color online) Three examples of networks sampled from a synthetic model with $N=36$ and $K_1=K_2=3$ are shown with $\lambda=\{0.0,0.5,1.0\}$, respectively. Two subnetworks are represented in an inner and outer ring of nodes with colors representing cluster assignments in each subnetwork. The darkness and width of edges represent edge weights.}\label{fig:ex-synth-networks}
\end{figure*}

\subsection{Overlapping networks for the stochastic blockmodels}
Under the above assumptions of Poisson observations any model for single networks that can be re-formulated to have Poisson observations can be used in a multi-network setting. This include diverse types of network models including for instance the hierarchical network model of \citeauthor{clauset2008hierarchical}~\cite{clauset2008hierarchical} or the overlapping community-type models such as \cite{miller2009nonparametric,latouche2011} which are easy to re-formulate as Poisson observations (trivially if one simple consider the Bernoulli probability as the rate in the Poisson model) or in the case of \cite{menon2010predicting} or \cite{PhysRevE.84.036103} already formulated in terms of Poisson observations, and one can also consider hybrids where different mixtures of models are used. 

However we will consider the simplest case where each network is modeled as a SBM. Many popular references exist on the SBM~\cite{holland1983stochastic,girvan2002community,kemp2006learning} however we will re-state it for completeness. The SBM assumes the $n$ vertices are divided into $\ell$ non-overlapping blocks. The assignment of vertex $i$ to block $\ell$ in network $s$ is indicated by $z_i^s = \ell$ and we denote by $\m z^s = (z_1^s, \dots, z_n^s)$ the assignment vector of all vertices in network $s$. As a prior for assignments we use the Chinese Restaurant Process (CRP) parameterized by a single parameter $\alpha$ controlling the distribution of group size~\cite{aldous1985exchangeability}, which we indicate by the symbol $\mathcal{C}$. Using our notation, this has a density given as
\begin{align}
 p(\m z^s|\alpha) &= \frac{\alpha^L\Gamma(\alpha)}{\Gamma(N+\alpha)}\prod_{\ell=1}^L\Gamma(n_\ell), \label{eq:crp}
\end{align}
where $\Gamma(\cdot)$ is the Gamma function and $\ell=1,\ldots,L$ indexes the blocks of network $s$. For further details on why the CRP is advantageous over, say a uniform prior, see \cite[p.2]{herlau2014idcsbm}. The generative process we then write up as:
\begin{subequations}\label{e1}
\begin{align}
& & \m z^s & \sim \mathcal{C}(\alpha), & & \nnspace \textit{clusters} \\
&\mbox{ for $\ell \leq m$ } \nnspace & \eta_{\ell m}^s & \sim \Gam(\kappa,\lambda), & & \nnspace \textit{link rate} \label{eq:sbm_poisson_link_rate} \\
& \mbox{ for $i < j$ } \nnspace & A_{ij}^s | \m z^s, \m \eta^s & \sim \Poisson(\eta_{z_iz_j}), & & \nnspace \textit{link weight} \label{eq:sbm_poisson}
\end{align}
\end{subequations}
In words, this process can be understood as follows
\begin{enumerate}[nosep,label=(\roman*)]
\item $\m z^s \sim \CRP(\alpha)$: Sample cluster assignments from the Chinese Restaurant Process \cite{aldous1985exchangeability} parametrized by a single parameter $\alpha$ controlling the distribution of group size, thus obtaining the partitioning of cluster associations for each vertex ($|\m z^s|=N$) into $1 \leq L \leq N$ clusters.
\item $\eta_{\ell m}^s \sim \Gam(\kappa,\lambda)$: Generate intra- and intercluster link rates from a Gamma distribution with shape parameter $\kappa$ and rate parameter $\lambda$.
\item $A_{ij}^s | \m z^s, \m \eta^s \sim \Poisson(\eta_{z_iz_j})$: Generate edges that are independently Poisson distributed with the expected number of links between vertices $i$ and $j$ being $\eta_{z_iz_j}$.
\end{enumerate}

An illustrative example of single-network and two-network MNSBM is shown in \figref{fig:example-networks}. Here the shaded regions in (b) and (c) surrounding the vertices represent blocks and next to the networks are given corresponding link rates. Both models \figref{fig:ex-disjoint-sbm} and \figref{fig:ex-mnsbm} will, given Poisson observation models, on average generate the observed network \figref{fig:ex-obs-network}.

\subsection{Inference and missing data\label{secs:method_inference}}
An efficient inference scheme is easily obtained through eqs.~\eqref{eq:9}, \eqref{eq:10} and \eqref{eq:11}. Notice when updating $\Phi^s$ in eq.~\eqref{eq:11} one can use the standard tool to integrate out the $\m \eta^s$ parameters.  This leaves only the assignments $\m z^s$ and hyperparameters $\alpha^s$, $\kappa^s$, $\lambda^s$ to be sampled. We sampled $\m z^s$ using standard Gibbs sampling and the hyperparameters using random-walk Metropolis-Hastings in log-transformed coordinates~\citep{kemp2006learning}. For simplicity we assumed a $\Gam(2,1)$ prior for $\alpha^s$, $\kappa^s$, $\lambda^s$. For details on deriving the collapsed Gibbs sampler, we refer to \cite{herlau2014idcsbm}.

To predict missing edges we used imputation. Suppose an edge $ij$ is unobserved. Then if we implement the sampler exactly as described but for this pair $ij$ replace eq.~\eqref{eq:9} with the unconstrained distribution
\begin{align}
p(A_{ij} | \cdots) = \mathcal{P}(A_{ij} | \eta_{ij})
\end{align}
we will get a sequence of MCMC estimates of $A_{ij}$, $(a_{ij}^{(t)})_{t=1}^T$. Predictions may then be estimated as the MCMC average
\begin{align}
p(A_{ij} = 1 | \m \Aobs) = \frac{1}{T}\sum_{t=1}^T h( a_{ij}^{(t)}).
\end{align}

In terms of computational complexity, a single Gibbs sweep over $\m z^s$ scales as $\bigO{EK^2}$, where $E=\sum_{ij} \m A^s_{ij}$, is the number of realized edges in $\m A^s$ (notice this is lower than the number of observed edges), and $K$ is the number of components. In addition the multinomial re-sampling steps in eq.~\eqref{eq:9},\eqref{eq:10} scales as $\bigO{E}$ taking up only a small fraction of the time spent. Thus with computational complexity $\bigO{EK^2}$ per Gibbs sweep. While this would indicate an computational cost roughly $S$ times greater than a single SBM it is worth emphasizing the stochastic de-coupling of networks in eq.~\eqref{eq:11} admits a very easy parallelization over Gibbs sweeps and with the advent of multi-core machines this allows the method to parallelize easily which we made use of. In addition in section \ref{section:sims} we will show the use of multiple networks allow each network to be modeled using fewer blocks reducing the quadratic factor in the cost. Taken together the cost of our method, when $S$ was lower than the number of cores on the machine, was comparable to the $S=1$ case. 
  




%% file: results.tex
\begin{figure*}[t]
	\centering
	\includegraphics[width=1\textwidth]{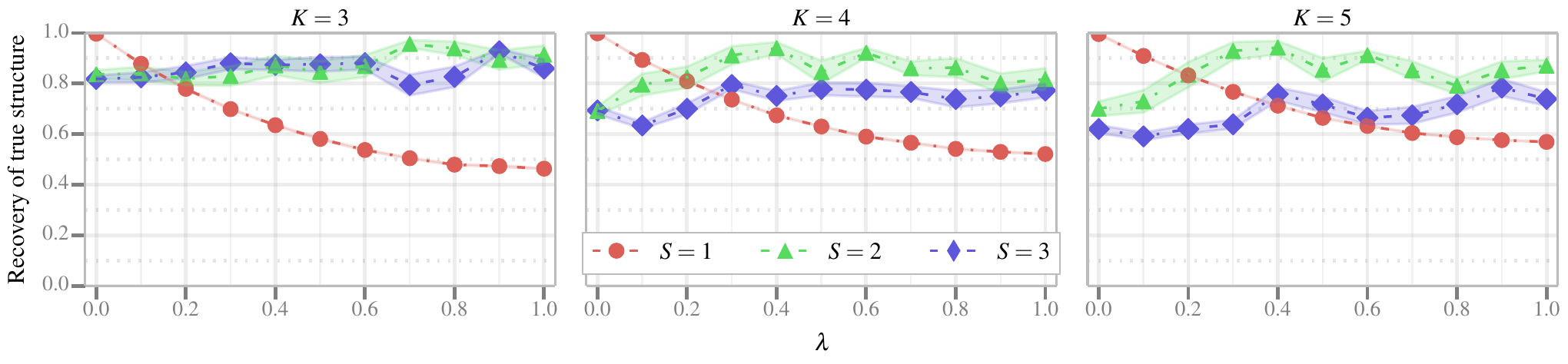}
	\caption{(Color online) The average AUC scores for each of $K=3,4,5$ and for $\est{S}=1,2,3$ with varying $\lambda$ for synthetic networks and ground truth being $S=2$. Each point is the average over 20 random restarts and shaded regions represent standard deviations of the mean. The experiments were run for $T=3000$ iterations and the AUC for each experiment is computed from the estimated same-block probabilities averaged over all Markov samples in the last 500 iterations.}
	\label{fig:synth-olp}
\end{figure*}

We test our method on both synthetic (computer generated) networks as well as a number of real world network datasets. Our synthetic benchmarks allow us to test the sampler as well as monitor how well the model identifies ground truth, i.e., planted structure, under controlled conditions. Analyzing real networks gives us an idea of the performance of our model in real-world scenarios.

\subsection{\label{sub:results_synthetic}Synthetic networks}

Since MNSBM is a generative model of networks, we can artificially sample data from known parameters. Running the model and comparing the solutions with ground truth, allows us to monitor how well the model performs when conditions are varied, e.g., community structure and area of overlap.

\subsubsection{Generating synthetic networks}
The synthetic data that we generate will serve to demonstrate that MNSBM can recover solutions with structure close to the ground truth.

We generate synthetic data by sampling from an instantiated MNSBM model with two subnetworks. In each subnetwork, we put $K_1=K_2=K$ equally sized clusters, i.e., the same number of communities for each subnetwork. We set the diagonal of $\m \eta^1$ to $1$, the diagonal of $\m \eta^2$ to $1.5$ and the off-diagonals of both to $0.1$ everywhere, thus generating strong community structure in both subnetworks. Setting slightly different link rates in the diagonals of $\m \eta^1$ and $\m \eta^2$ helps with the identifiability of the true structure. The overlap in the generated networks we control by circular shifting the clusters in the second subnetwork. Hence, in a network of $N$ nodes and a circular shift of $m$, the number of overlapping nodes is $mN/K$, where for notational simplicity we assume $N$ is a multiple of $K$. Consequently, with a shift $m>0$, there are $N-mN/K$ nodes in non-overlapping clusters as well as $mN/K$ nodes in overlapping clusters. Since we control overlap by a circular shift of the clusters in one subnetwork, the structure is trivially symmetric around $m=N/(2K)$, i.e., structurally there is no difference whether we shift $m=N/(2K)-i$ or $m=N/(2K)+i$ for all $i\leq N/(2K)$. Therefore we are interested in varying the overlap $m=0,1,\ldots,N/(2K)$ and we define a parameter $\lambda = 2Km / N$ as a discrete scale between 0 and 1, measuring from no shift up to the maximum $N/(2K)$. In \figref{fig:ex-synth-networks} we show an example of synthetic networks generated as described above using $N=30$, $K=3$, and $\lambda=\{0.2, 0.6, 1.0\}$.

Using $K=\{3,4,5\}$, we generate networks with $N=\{60,80,100\}$ nodes, respectively, and vary $\lambda$ in the interval $(0,1)$.

\begin{figure*}[!t]
	\centering
	\includegraphics[width=1\textwidth]{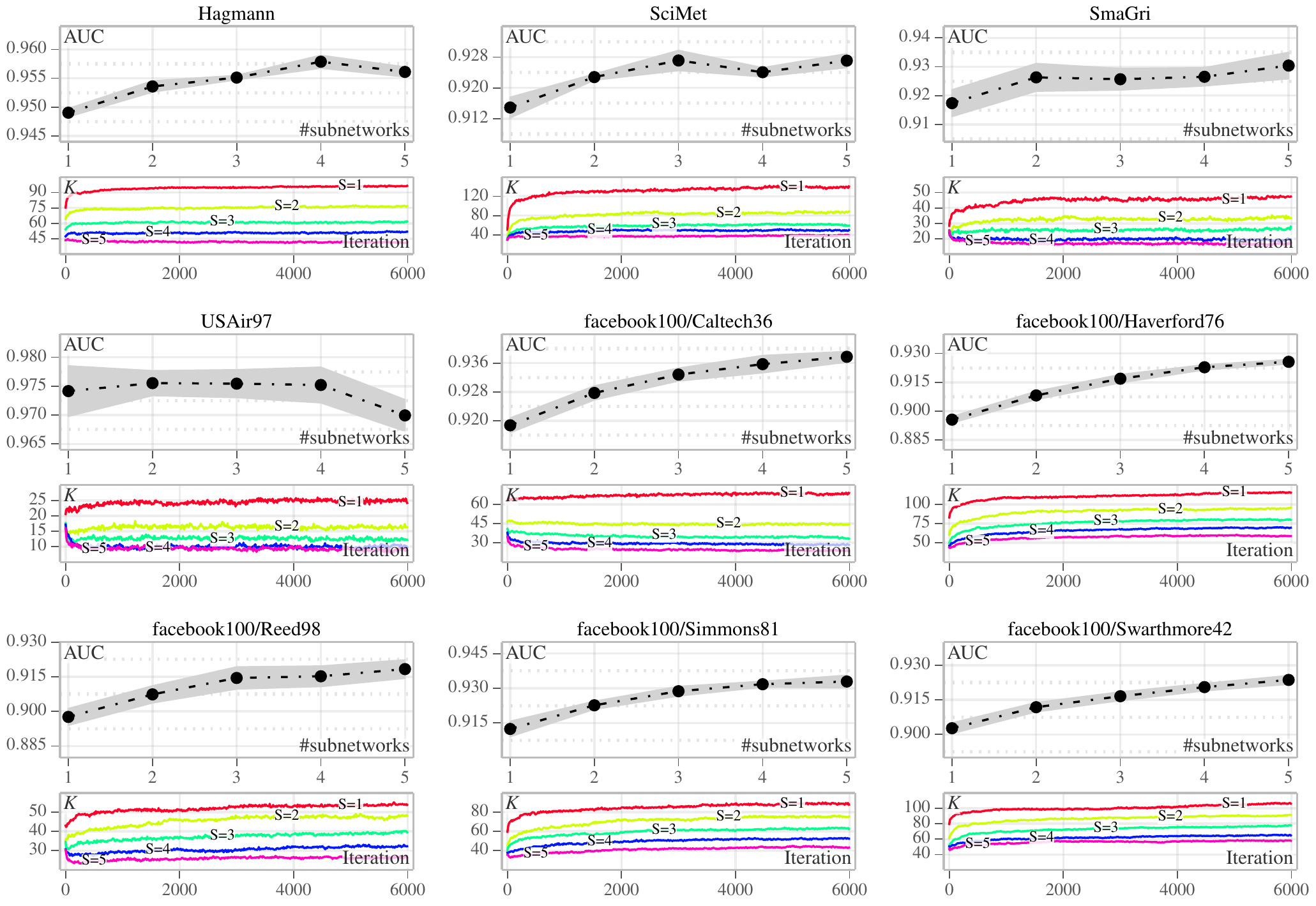}
	\caption{(Color online) MNSBM results for each of the networks introduced in the text. For each network, MNSBM is run with $S=\{1,2,3,4,5\}$, i.e., varying number of subnetworks, and for 6,000 iterations. In the upper plots the average AUC link prediction scores are shown as a function of $S$. The AUC score for a single chain is computed as the average of the predictions from every 10$^{th}$ iteration of the last half of the chain, discarding the first 3,000 as burn-in. The averages are based on 5 random restarts and 5\% edges and non-edges picked randomly for testing. Shaded regions represent the standard deviation of the mean. In the lower plots we show the average number of detected components per subnetwork as a function of iterations, with each line representing an MNSBM with a different $S$. The averages are computed similarly to the AUC scores.}
	\label{fig:networks-both}
\end{figure*}

\subsubsection{Performance on synthetic networks}

When testing our model on synthetic data, we are interested in how well our inference procedure is able to identify planted structure. In order to measure similarity between true and estimated models, however, we identify that similarity measures on the true and estimated assignment vectors directly introduces a matching problem. To circumvent this problem, we opt instead 
for a measure of match between the overall structures as follows. Suppose we are trying to infer the block structure of an artificially constructed graph, $\bm A^*$, containing $S^*$ true subnetworks using a MNSBM with $S>S^*$ subnetworks. Then, assuming the MNSBM works correctly, one of the subnetworks will be empty and the community structure will not be informative. To avoid empty subnetworks to influence our results, we will therefore focus on whether the MNSBM partitions the realized edges correctly. This can be done by, for each edge $e_k=(i,j)$ of $\bm A^*$, determine (i) if the particular true subnetwork which generated $\bm A^*$, $\bm A^{*s}$, assigned $(i,j)$ to the same block and (ii) compare this to whether the subnetwork $A^s$ in the inferred structure which "explains" $(i,j)$ (i.e.~has $A^s_{ij} = 1$) also assigns $(i,j)$ to the same block. Since each true edge may be explained by multiple subnetworks, we simply compute the weighted average over each subnetwork explaining this edge. This results in a binary vector of $a_k$'s consisting of the true same-block information and a weighted vector of $w_k$'s consisting of the estimated same-block probability averaged over all $T$ Markov samples. Specifically, for any given edge $e_k = (i,j)$ we define: 
\begin{align}
	a_k &= \sum_{s=1}^S A^{*s}_{ij}\delta_{z^{*s}_i=z^{*s}_j},\\
	w_k &= \frac{1}{\sum_{s}^S A^s_{ij}} \sum_{s=1}^S A^s_{ij} \delta_{z^s_i = z^s_j}
\end{align}
where $\delta_{(\cdot)}$ is an indicator function that is 1 if the condition ($\cdot$) is true and 0 otherwise. We can then compute area under the curve (AUC) of the receiver operating characteristics based on $a_k$ and $w_k$.

In \figref{fig:synth-olp} we show the dependencies of $K$ (plot titles), the choice of $\est{S}$ (different curves) and the overlap parameter $\lambda$. We observe that for $\lambda=0$, the simplest network model is always best, which is unsurprising since there is no overlapping structure. Since the single network, $\est{S}=1$, infers disjoint clusters for the overlapping structure, which increases in size as $\lambda \rightarrow 1$, we see a decline in similarity. For $\lambda > 0.2$ we begin to see the model with two subnetworks, $\est{S}=2$, outperforming the single network model. With the networks sizes that we have chosen for these experiments, $\lambda=0.1$ means there is only one node in each overlap and it looks as if it is necessary with a bit more nodes per overlap, i.e., three ($\lambda=0.3$), in order for the $\est{S}=2$ model to consistently pick up the structure. 

For $\est{S}=3$, where the model allows too many subnetworks, we see the similarity degrades compared to using $\est{S}=2$. I.e., the sampler is not perfectly able to set to zero all the edges of one of the subnetworks. As we will see in our experiments on real networks, choosing \est{S} too large does however not necessarily mean worse predictive performance. In practice one seldom knows the underlying structure, so for empirically assessing which \est{S} is better, we must resort to other strategies. In that case, we suggest cross-validating \est{S} on a held-out sample of edges w.r.t.~a problem specific measure such as AUC, RMSE, MAE, etc., depending on the task at hand, and picking $\est{S}_{\rm opt}$ based on a plot of the target measure as a function of $\est{S}$.

\subsection{Real networks}\label{section:sims}
We have tested MNSBM on a number of real world network datasets, where we are interested in \textit{link prediction} for out-of-sample edges. We vary the number of sub-networks $S$ from 1 to 5 and repeatedly run inferences five times with random initialization and a random subset of 5\% edges (and a similar number of non-edges) held-out for measuring link prediction performance. The link prediction performance we report in area under curve (AUC) of the receiver operating characteristics.


\begin{figure}[!t]
	\centering
	\includegraphics[width=0.9\columnwidth]{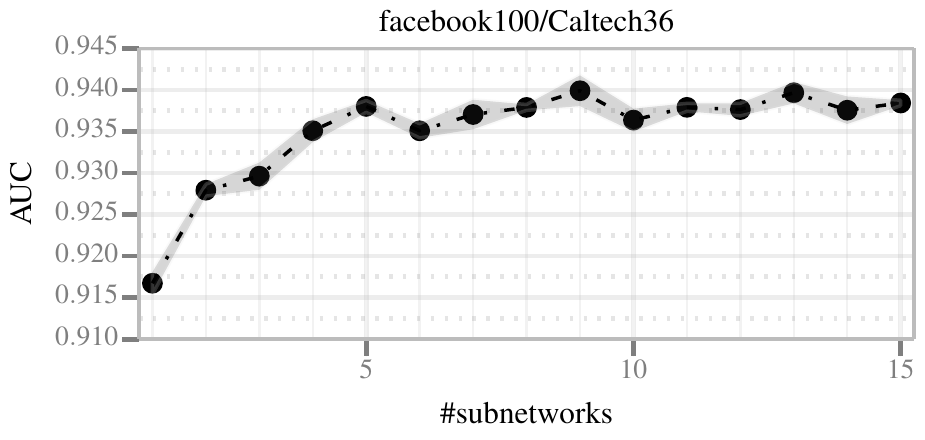}
	\caption{AUC score as a function of the number of subnetworks for the \emph{facebook100/Caltech36} network. The plotted values are averages and the shaded region represents standard deviation of the mean based on the same experimental settings as in \figref{fig:networks-both}.}
	\label{fig:caltec-auc-bigs}
\end{figure}

The nine networks we analyze are.
\begin{enumerate}[nosep,label=(\roman*)]
	\item \emph{Hagmann:} undirected weighted network of the number of links between 998 brain regions as estimated by tractography from diffusion spectrum imaging across five subjects \cite{hagmann2008mapping}. As in \cite{herlau2014idcsbm}, the graph of each subject has been symmetrized, thresholded at zero and the five subject graphs added together.
	\item \emph{SciMet:} directed weighted network of a citation network between 3,086 authors in the Scientometrics journal 1978-2000 \cite{pajek2006datasets}.
	\item \emph{SmaGri:} directed weighted network of another citation network between 1,059 authors to Small \& Griffith and Descendants \cite{pajek2006datasets}.
	\item \emph{USAir97:} undirected weighted network of air traffic flow between 332 U.S.~airports in 1997 \cite{neal2010refining, neal2014airnet, pajek2006datasets}.
	\item \emph{facebook100/*:} undirected unweighted networks from five friendship networks in U.S.~colleges from the Facebook100 dataset \cite{facebooklong}. \emph{Caltech36:} 769 nodes, \emph{Haverford76:} 1446 nodes, \emph{Reed98:} 962 nodes, \emph{Simmons81:} 1518 nodes, \emph{Swarthmore42:} 1659 nodes.
\end{enumerate}

For all the directed networks, we symmetrize them, i.e., treat them as undirected, and for the weighted networks, we make them unweighted by treating any non-zero edge as a link.

The results are shown in \figref{fig:networks-both}. For each of the networks, we show in the top the average AUC scores with the shaded region being standard deviations of the average based on five random restarts and different held-out samples. In the bottom, we show the evolution of average number of inferred communities per subnetwork. As we increase $S$, the number of subnetworks in MNSBM, we generally see the AUC scores increasing, meaning that the link prediction for missing edges improves. The only exception is \emph{USAir97}, which is also the smallest network in our benchmark. Either \emph{USAir97} does not exhibit overlapping structure or the small size of the dataset is an inhibiting factor. In terms of average number of inferred communities, we see that this is consistently decreasing as we increase $S$ and eventually saturates. These experiments confirm that on a variety of real world networks, MNSBM enables modeling ensembles of simpler substructures while increasing link prediction performance.

As a separate experiment, we have run additional five independent trials with the \emph{facebook100/Caltech36} network, where we let $S$ increase to 15. The AUC as a function of $S$ is shown in \figref{fig:caltec-auc-bigs}. We see that after $S=5$ the performance saturates, which shows that our method is robust towards choosing too high number of subnetworks. I.e., overfitting is not really a concern here. In practice what happens is that  as the sampler progresses, superfluous subnetworks get very few edges assigned (if any), hence they become redundant without affecting the performance negatively.

%% file: conclusion.tex
In this work we have discussed two views on extending existing methods for structural network modeling. In the \textit{communities first} view, which we argue is prominent in recent complex networks research, the (simplifying) assumption of structural equivalence is sacrificed in order to allow for overlapping groups leading to evermore complicated definitions on what constitutes a "group" and how these groups interact. We explore an alternative view which we attribute to the seminal work of \citeauthor{white1976social}, which we dub the \textit{networks first} view. The key distinction is that it considers the complexity of observed networks as arising as a consequence of multiple networks of different types of ties (edges) being aggregated. Inspired by the latter view of complex networks we introduce the multiple-networks stochastic blockmodel (MNSBM), which seeks to \textit{separate} the observed network into subnetworks of different types and where the problem of inferring structure in each subnetwork can benefit from the simplifying assumption of structural equivalence. The result is effectively the joint inference problem of splitting the observed edges between subnetworks and identifying (block) structure in each subnetwork. We formulate this model in a generative Bayesian framework over parameters that can be inferred efficiently using Gibbs sampling. Thereby we obtain an effective multiple-membership model without introducing the drawbacks that originate from defining complex interactions between groups. We demonstrate results on the recovery of planted structure in synthetic networks, as well as provide results in terms of link prediction performances on a number of real-world network data sets, which highly motivate the use of multiple subnetworks over a naive stochastic blockmodel, assuming disjoint blocks globally on the networks.